\documentclass[sigconf]{acmart}

\renewcommand\footnotetextcopyrightpermission[1]{} 
\usepackage{booktabs} 
\usepackage{algorithm}
\usepackage{algpseudocode}
\usepackage{float}
\usepackage{caption}

\setcopyright{none}

\begin{document}
\title{ Semantic See-Through Rendering on Light Fields }

\author{Huangjie Yu}
\authornote{ShanghaiTech University, Shanghai, China\newline
Email: [yuhj, zhanggl, mayx, zhangyl, yujingyi]@shanghaitech.edu.cn}
\affiliation{
}

\author{Guli Zhang}
\authornotemark[1]
\affiliation{
}

\author{Yuanxi Ma}
\authornotemark[1]
\affiliation{
}

\author{Yingliang Zhang}
\authornotemark[1]
\affiliation{
}

\author{Jingyi Yu}
\authornotemark[1]
\affiliation{
}

\begin{abstract}
    We present a novel semantic light field (LF) refocusing technique that can achieve unprecedented see-through quality. Different from prior art, our semantic see-through (SST) differentiates rays in their semantic meaning and depth. Specifically, we combine deep learning and stereo matching to provide each ray a semantic label. We then design tailored weighting schemes for blending the rays. Although simple, our solution can effectively remove foreground residues when focusing on the background. At the same time, SST maintains smooth transitions in varying focal depths. Comprehensive experiments on synthetic and new real indoor and outdoor datasets demonstrate the effectiveness and usefulness of our technique.
\end{abstract}

%
%



\begin{teaserfigure}
  \includegraphics[width=\textwidth]{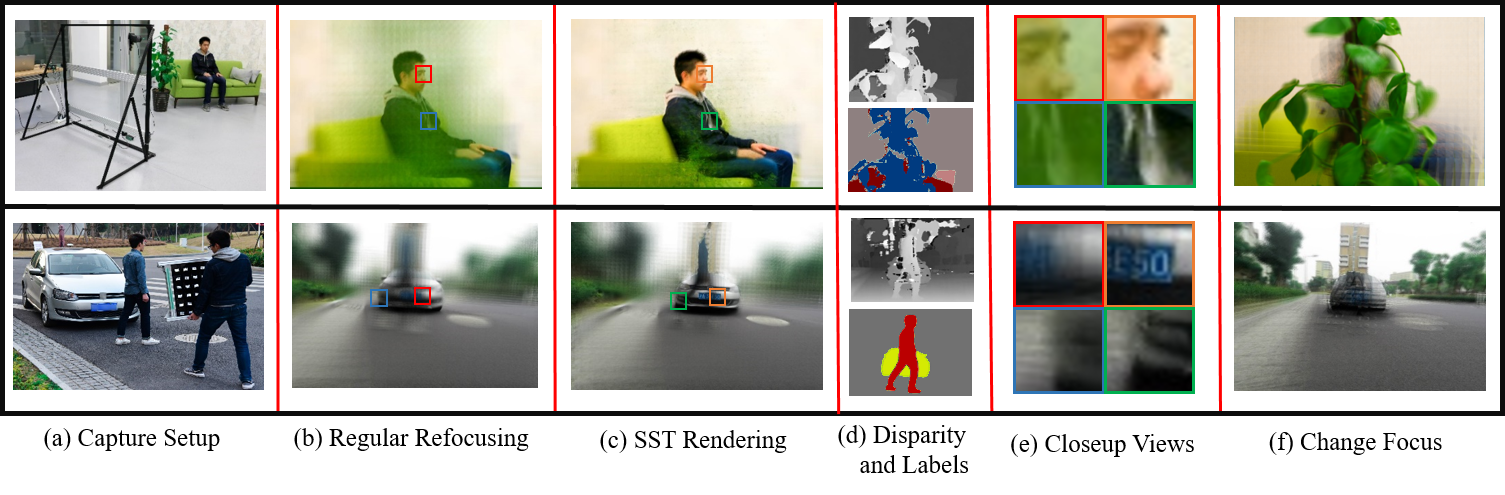}
  \caption{Our semantic see-through (SST) vs. regular LF refocusing. Top row: results on the "bush" LF captured by a motor rig. When focusing on the asset (person), SST (c) manages to remove nearly all foreground bush whereas regular refocusing (b) exhibits strong foreground residue. Bottom row: results on the "pedestrian" LF captured by a GoPro camera array; SST manages to reveal the plate number when focusing on the car (c) while regular refocusing cannot (b). (d) shows the corresponding depth map and label results of the central LF view. (e) shows the closeup views. (f) shows SST refocusing at a different depth.}
  \label{fig:Teaser}
\end{teaserfigure}

\maketitle

\section{Introduction}
A plenoptic function describes rays originating from any point and along any direction in free space~\cite{Adelson91theplenoptic}. A practical implementation of the plenoptic function is the light field~\cite{Levoy1996LightFR}, widely used for image-based modeling and rendering and most recently for virtual and augmented reality~\cite{lytro_2018,otoy_2018}. A LF adopts the two plane parametrization (2PP): each ray is parameterized by its intersections of two parallel planes $st$ and $uv$ at $z = 0$ and $z = 1$ respectively as a 4-tuple $r = [s, t, u, v]$. Under the camera array setting, $[s, t]$ can be viewed the camera position and $[u, v]$ as the pixel coordinate in the respective camera ~\cite{Wilburn2004HighspeedVU, Yang2002ARD}.

A unique rendering capability of LF is post-capture refocusing ~\cite{Isaksen2000DynamicallyRL, Ng2005FourierSP}: given a virtual focal plane $z = d_f$, rays from an LF are resembled as if they emit from the focal plane:
\begin{equation}
E(u' , v') = \iint
L(s,t,s+\frac{u'-s}{d_f},t+\frac{v'-t}{d_f})A(s,t)\mathrm{d}s\mathrm{d}t
\end{equation}
where $A$ represents the virtual aperture that controls the angular extent of rays to gather. The refocusing process can be further accelerated in the frequency space~\cite{Ng2005LightFP}. More recent approaches further employed depth-guided ray interpolation to minimize visual artifacts such as aliasing~\cite{Yang2016} and color bleeding~\cite{Fiss_2015_CVPR}, where the depth map can be acquired either by active 3D sensing~\cite{6162880} or passive light field stereo matching~\cite{Wanner2012GloballyCD, jeon:cvpr15}.

By setting the aperture really big (e.g., using a LF camera array), the refocusing effect can further mimic virtual see-through~\cite{Wilburn2005HighPI, Isaksen2000DynamicallyRL}. Fig.\ref{fig:Teaser} (b) shows that by focusing on the asset hiding behind the bushes, we can partially remove the foreground. However, even at a full aperture (i.e., using all cameras in the array), the rendering still exhibits substantial foreground residue.

\begin{figure*}
\includegraphics[width=1\textwidth]{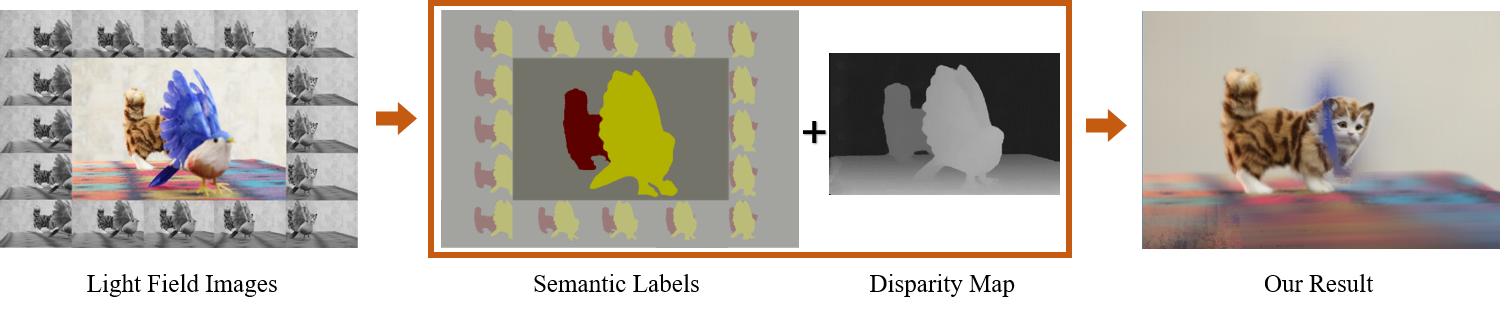}
\caption{Our Semantic See-Through Rendering Pipeline: We obtain initial semantic labels using deep learning, refine the results using stereo matching, and conduct semantic-based refocusing.}
\label{fig.rendering}
\end{figure*}

In this paper, we introduce a novel LF see-through rendering technique that we call Semantic See-Through or SST. In a nutshell, SST differentiates rays emitting from objects of different semantic meanings (labels) as well as at different depths. SST then encodes such differences into the blending function by assigning a weight to each ray $r$. Fig.\ref{fig.rendering_result_1} shows result of SST vs. regular refocusing on the same bush scene. Using the weighting scheme described in Sec.\ref{sec:weighting}, SST achieves superior see-through quality by eliminating nearly all foreground bushes. At the same time, it maintains shallow depth-of-field (DoF) effect when changing focus as shown in Fig.\ref{fig.rendering_result_1} and the supplementary video.  

We first present a light field semantic labeling scheme based on deep neural networks. Our technique builds upon the Pyramid Scene Parsing Network (PSPNet)~\cite{8100143} that employs contextual relationship and global information can handle complex scenes. We extend the PSPNet framework to output a probability map for each light field view as the raw semantic map and refine the results using the LF stereo matching techniques. Specifically, we conduct learning-based stereo matching and employ warping to compensate for occlusions. This allows us to assign each ray a depth and a semantic label. Finally we develop a ray weighting scheme based on depth, semantic label, and focal depth. Comprehensive experiments on synthetic and new real-world datasets demonstrate that our technique, although simple, produces unprecedented see-through quality.

\section{Related work}
Our work is closely related to latest advances in semantic segmentation and multi-view reconstruction. Due to the space limit, we only discuss the most relevant ones.

\paragraph{Light Field Rendering.} Shortly after Adelson's introduction of the Plenoptic function~\cite{Adelson91theplenoptic}, Levoy and Hanrahan ~\cite{Levoy1996LightFR} proposed the  LF model as a practical representation to the Plenoptic function. This 4D representation enables direct ray space interpolation for new view synthesis and is most recently employed in VR content creation~\cite{otoy_2018}. In a similar vein, Lumigraph uses irregularly sampled views while guiding ray interpolation with geometry proxies. Chai et al.~\cite{Chai344932} analyzed the relationship between aliasing and sampling. 
\cite{Isaksen2000DynamicallyRL} demonstrated refocusing on LFs by blending rays passing through an aperture and converging on a focal plane. By changing the depth of the focal plane, they can produce a focal stack. The process is further accelerated using Fourier space slicing ~\cite{Ng2005FourierSP} on datasets captured by plenoptic cameras such as Lytro~\cite{lytro_2018}. All previous approaches treat rays "equal", discarding their heterogeneity in depths or semantic meanings.

\paragraph{3D Reconstruction.} The special sampling pattern in LFs has also renewed interest on stereo matching~\cite{7432007, Heber2016ConvolutionalNF,
Huang_2017_ICCV, SAG17:cvpr, 7951484}. Tao et al. combined different appearance cues to estimate dense depth maps~\cite{tao2015shading}. Chen et al.~\cite{Chen2014LightFS} characterized the bilateral consistency on angular ray patches. Wang et al. employed an edge detector and analyzed pixel-wise occlusion along the edges to improve quality near occlusions~\cite{wang2015occlusion}. Lin et al.~\cite{Lin2015DepthRF} discovered the focal symmetry property and applied an analysis-by-synthesis technique to handle noise and occlusions. Zhang et al.~\cite{Zhang_2017_ICCV} suggested using ray-space structure-from-motion for fusing stereo matching results from multiple LFs. These techniques are suitable for LFs of small baselines (e.g., the Lytro data). For see-through effects, it is desirable to use LFs of very large baselines.  

\paragraph{Semantic Labeling.} Latest semantic labeling techniques exploit rich data and deep learning techniques~\cite{he2017maskrcnn, CP2016Deeplab, 8100143}. The availability of semantic labels has further benefited many traditionally challenging computer vision problems, ranging from stereo matching and volumetric
reconstruction~\cite{Ladicky2010JointOF, Tulsiani_2017_CVPR} to tracking~\cite{7487374, 6619046, 7487378}. By far, nearly existing techniques are developed for 2D images rather than a 4D light field. Wanner et al.~\cite{Wanner2012GloballyCD, Wanner-et-al-2013} presented a coherent light field segmentation technique although their technique does not
explicitly use semantic information. We seek to assign each ray in the light field with a semantic label and then employ the results to improve the see-through effects. The seminal work of Pyramid Scene Parsing Network (PSPNet) discovers contextual relationship and global information in complex scenes. The results are outstanding on popular benchmarks. However, their results are sensitive to occlusions and brute-force application to individual LF views leads to incoherence. 

\paragraph{Semantic-Guided Reconstruction.} Finally, our technique also aligns with recent trends on incorporating semantic information in 3D reconstruction. The seminal work by Kundu et al. used a Conditional Random Field (CRF) model to jointly estimate object labeling and voxels occupation~\cite{Kundu2014JointSS}. H\"ane et al. further proposed a joint optimization framework to simultaneously solve for the image segmentation and 3D reconstruction problems. The key observation is that semantic segmentation and geometric cues are complementary~\cite{6618864, 7575643}. Although effective, the quality of volumetric reconstruction relies heavily on the voxel resolution. Conceptually, a LF can also be viewed as a multi-view representation and we can directly use volumetric semantic labeling to label each ray. In reality, treating every ray as a voxel is computationally infeasible due to high resolutions of a LF. 

\section{Semantic Light Field Labeling}

We extend the pyramid scene parsing network (PSPNet)~\cite{8100143} to incorporate multi-scale contextual information in pixel-wise label prediction. To coherently conduct semantic labeling to all views in a LF, we design a two-stage deep network pipeline. The first stage employs a PSPNet to generate an initial semantic label map as well as the marginal distributions of every pixel label. Our second stage introduces a high confidence semantic map (HCSM) prediction module that estimates how reliable each pixel label prediction is. The HCSM is then used to filter out unreliable pixels with low confidence values. Here we first briefly review the PSPNet architecture used in our work and then discuss the HCSM generation module. 

\begin{figure*}
\includegraphics[width=1\textwidth]{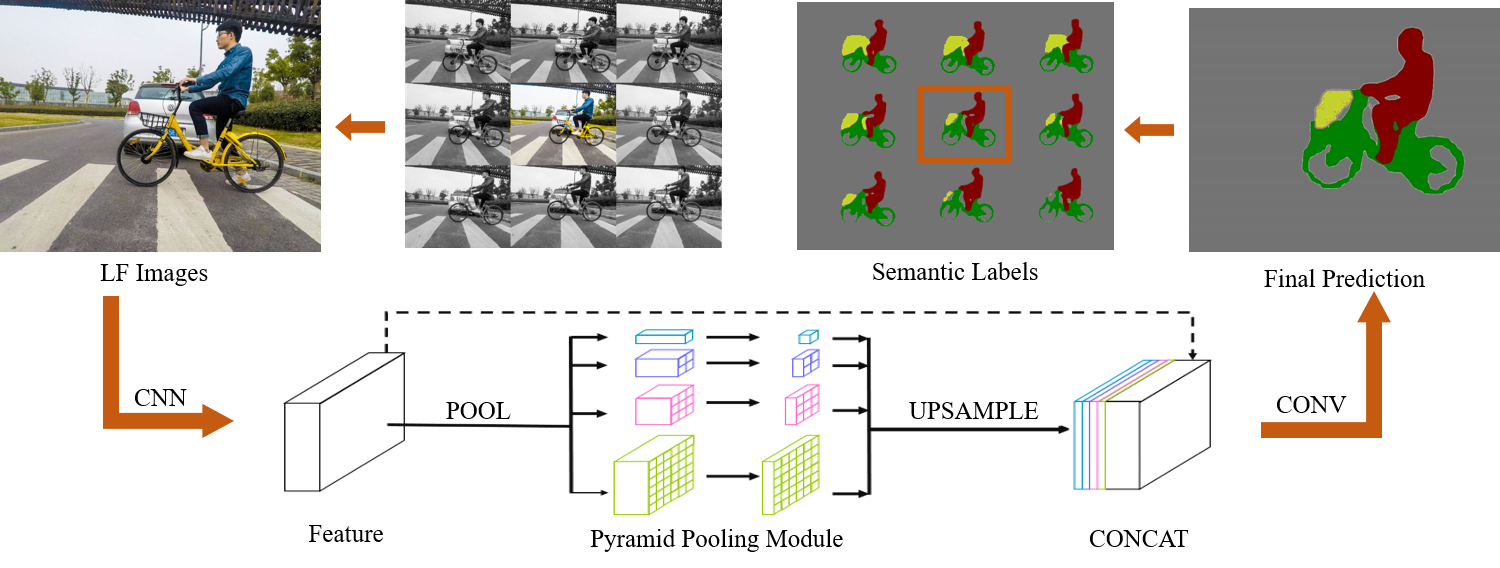}
\caption{Initial Semantic Labeling: We apply the Pyramid Scene Parsing (bottom row) to label individual LF views.}
\label{fig:PSPNet}
\end{figure*}

The PSPNet first computes a convolutional feature map based on the ResNet and dilated convolutions, which has $\frac{1}{8}$ of the input image size. To encode spatial context information, the network adopts a pyramid feature pooling strategy that generates a four-level feature maps representing global and sub-regional contexts. The pooled context feature maps are then upsampled and concatenated with the original feature map as inputs to predicting a multi-class label distribution for each pixel. The final semantic labeling can be generated by taking the most-probable label configuration for the entire image. The model configuration is illustrated in Fig.\ref{fig:PSPNet}.

Formally, we denote the label space as $\mathcal{C}=\{1,\cdots, C\}$ where $C$ is the number of semantic classes. At each pixel $x_i$, which corresponds to a visual ray $[u,v,s,t]$, the PSPNet computes a label distribution $p(y_i)$ where $y_i\in\mathcal{C}$. We generate the pixel-wise labeling by taking the MAP estimates as $y_i^*=\arg\max p(y_i)$.      

Given the outputs of the PSPNet, we now estimate a confidence score for each pixel's label prediction. We observe that the marginal label distribution $p(y_i)$ tends to be more divergent in complex areas, e.g., near the boundary of an object, and more concentrated within an object region. We therefore compute a HCSM in terms of the entropy of the label distribution as follows,
\begin{equation}
H(x_i) = -\sum_{y_i\in\mathcal{C}} p(y_i)\log p(y_i)
\end{equation}
The label distribution at each pixel reaches the maximum entropy when each label value shares the same probability, and reaches the minimum when it takes a single label value with probability $1$. A large entropy generally indicates that we are not confident at that pixel due to diverse label probabilities. 

Given the HCSM, we use a simple thresholding strategy to filter out pixels with unreliable label predictions. Specifically, we consider the initial semantic label prediction $y_i^*$ confident if the following condition holds,
\begin{equation}
H(x_i) = -\sum_{y_i\in\mathcal{C}} p(y_i)\log p(y_i) < \epsilon_H
\end{equation}
where $\epsilon_H$ is a model parameter controlling the balance between the precision and recall of the remaining pixel labeling. 

To estimate the parameter $\epsilon_H$, we introduce a score function based on the quality of the label predictions satisfying the above condition. Denote the remaining pixel set as $S_\epsilon$, we first use the semantic label map $Y^*_{S_\epsilon}$ to estimate its accuracy $Acc = \frac{TP}{TP + FP}$ (eliminating background label) and coverage $Cvg = \frac{TP}{TP + FN}$ (eliminating background label). (We manually label one view to calculate $Acc$ and $Cvg$) A larger $\epsilon_H$ usually indicates low accuracy but high coverage and vice versa. To achieve a balance between accuracy and coverage, we estimate $\epsilon_H$ by maximizing the following score function:
\begin{equation}
Score = Acc ^m \cdot Cvg
\end{equation}
where $m$ is a hyper-parameter indicating the importance of accuracy over coverage. A higher $m$ tends to output more accurate semantic map. In this work, we choose $m=4$.

By applying the confidence-based thresholding, we can remove large amount of unreliable label predictions and improve the precision of the semantic maps in most of cases. In general, the filtered label maps are more accurate than the initial semantic maps that directly comes from the convolutional neural network. However, due to the filtering, each image may have some regions with missing label predictions. We can fill in semantic labels for these missing regions using stereo refinement.

\section{Stereo-based Semantic Refinement}
For real-world scenes, PSPNet fails in multiple cases. When object B occludes object A, PSPNet may label the fragments neither label A or B. Further since PSPNet assumes a default minimal object size, small sized objects cannot be properly labeled either. Finally, generating a HCSM further decreases label coverage. The failure can cause serious visual artifacts. Our approach is to combine the HCSM with the disparity maps: an object, even fragmented by occlusions, generally has coherent depths. We employ such depth coherent to improve the semantic labeling quality as shown in Fig.\ref{fig.label_refinement_result}.

\subsection{Disparity Map Generation}

There are a number of LF stereo matching techniques publicly available. However, nearly all of these approaches are tailored to handle LFs with ultra-small baselines such as the ones captured by the plenoptic camera. Since meaningful see-through effects require a wide synthetic aperture and inherently a large baseline between LF views, we resort to pair-wise stereo matching and then refine the results via warping. 

Specifically, we employ a learning-based MC-CNN by Zbontar et al.~\cite{zbontar2016stereo} to first conduct pair-wise disparity map estimation. MC-CNN uses the convolutional neural network to initialize the stereo matching cost:
\begin{equation}
C(\textbf{p}, d) = -s(P^L(\textbf{p}), P^R(\textbf{p} - \textbf{d}))
\end{equation}
where $P^L(\textbf{p})$ is a patch from left image, $d$ is the disparity under consideration and  $P^R(\textbf{p} - \textbf{d})$ is a patch from right image. $s(P^L(\textbf{p}), P^R(\textbf{p} - \textbf{d}))$ is the output from the neural network which indicates the similarity between the two patches.

MC-CNN iteratively applies cross-based cost aggregation to average matching cost over a support region. It differs from averaging in a fixed window in that pixels in a support region belong to the same physical object. It then imposes smoothness on the disparity image by minimizing the following energy function in two horizontal and two vertical directions:
\begin{equation}
\begin{split}
E(D)=\displaystyle\sum_\textbf{p} \{C(\textbf{p}, D(\textbf{p})) + \displaystyle\sum_{\textbf{q}\in\mathcal{N}_\textbf{p}} P_1 \cdot 1(|D(\textbf{p}) - D(\textbf{q})|=1) +\\
\displaystyle\sum_{\textbf{q}\in\mathcal{N}_\textbf{p}} P_2 \cdot 1(|D(\textbf{p}) - D(\textbf{q})|>1)\}
\end{split}
\end{equation}
where $1(\cdot)$ is the indicator function, $P_1$ are $P_2$ are smoothness penalties.

After sub-pixel enhancement, MC-CNN finally refines the disparity map by introducing a $5\times5$ median filter and the following bilateral filter:
\begin{equation}
\begin{split}
D_{Final}(\textbf{p}) = \frac{1}{M}\displaystyle\sum_{\textbf{q}\in\mathcal{N}_\textbf{p}} D(\textbf{q}) \cdot g(\|\textbf{p}-\textbf{q}\|) \cdot 1(|I(\textbf{p})-I(\textbf{q})| < \epsilon_I)
\end{split}
\end{equation}
where $g(x)$ is the standard normal distribution, $\epsilon_I$ is the intensity blurring threshold and $M$ is the normalizing constant.

MC-CNN has shown superior performance over the state-of-the-art. However, directly using those disparities maps causes problems. This is mainly because the disparity map from a pair of images contains holes, especially on the left margin of the left image and the right margin of the right image. 

Recall that an LF is composed of a number of rectified image pairs. This provides sufficient number of disparity maps to patch in the holes by warping and interpolating individual disparity map. Specifically, we generalize the interpolation scheme in ~\cite{zbontar2016stereo} to multiple views. 

Let $D^R$ denote the reference disparity map in which we hope to patch the holes. Let $\{D^{L_1}, D^{L_2}\}$ denote the left two disparity maps of the reference disparity map. An incorrect disparity pixel in $D^R$ represents inconsistency between $D^R$ and $D^L$. Therefore, we label each disparity $d$ in $D^R$ by performing the following consistency check on $D^R$ and $D^L$ : 

\[ label = 
  \begin{cases}
    \textbf{correct}        & \quad \text{if } |d - D^{L_1}(\textbf{p}+\textbf{d})| \leq 1 \text{ or } \\
                  & \quad \;\;\; |d - D^{L_2}(\textbf{p}+\textbf{2d})| \leq 1 \text{ for } d=D^R(\textbf{p})\\
    \textbf{mismatch}  & \quad \text{if } |d - D^{L_1}(\textbf{p}+\textbf{d})| \leq 1 \text{ or } \\
                  & \quad \;\;\; |d - D^{L_2}(\textbf{p}+\textbf{2d})| \leq 1 \text{ for any other } d\\
    \textbf{occlusion}  & \quad \text{otherwise}
  \end{cases}
\]

\begin{figure}
\includegraphics[width=1\linewidth]{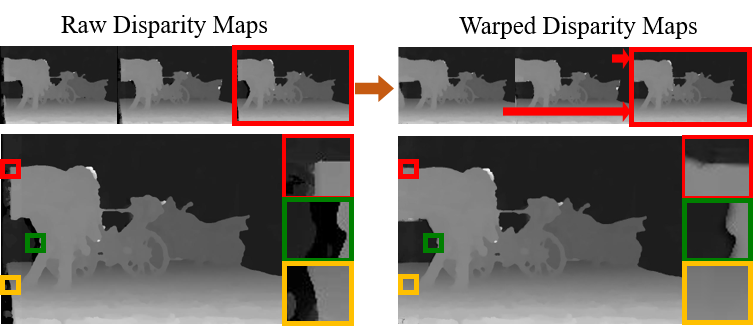}
\caption{Handling Occlusions in Stereo Matching. Left: raw pairwise stereo matching results; Right: we warp the stereo results onto individual views to patch holes. }
\label{fig:depth_warp}
\end{figure}

For position $\textbf{p}$ marked as \textbf{occlusion}, we perform a linear right search in $D^{L_1}$ until we find a \textbf{correct} position $\textbf{p'}$ that satisfies $\textbf{p'}-D^{L_1}(\textbf{p'})=\textbf{p}$. If the search fails, we further search in $D^{L_2}$ until we find a \textbf{correct} position $\textbf{p''}$ that satisfies $\textbf{p''}-D^{L_2}(\textbf{p''})=\textbf{p}$. For position marked as \textbf{mismatch}, we apply the same interpolation strategy as in ~\cite{zbontar2016stereo}. We also consider the left margins in each view as occlusions and perform a linear search as above. The resulting disparity maps are illustrated in Fig. \ref{fig:depth_warp}.

\subsection{Disparity-Guided Semantic Refinement}
To use disparity maps to refine semantic labeling, we assume that objects of different labels should have different depths. Notice if there are multiple objects lying at the same depth, they clearly will not occlude each other and therefore our refinement will not affect the final result. Next, we fit a normal distribution $g_i(d)$ with respect to disparities of every label in $\mathcal{C}$, where $d$ is disparity. It then becomes straightforward to label each pixel according to its disparity value, i.e., we assign label $i$ to pixel $[u, v, s, t]$ if $g_i(D(u, v, s, t)) > \epsilon_D$. 

\begin{figure*}
\includegraphics[width=1\textwidth]{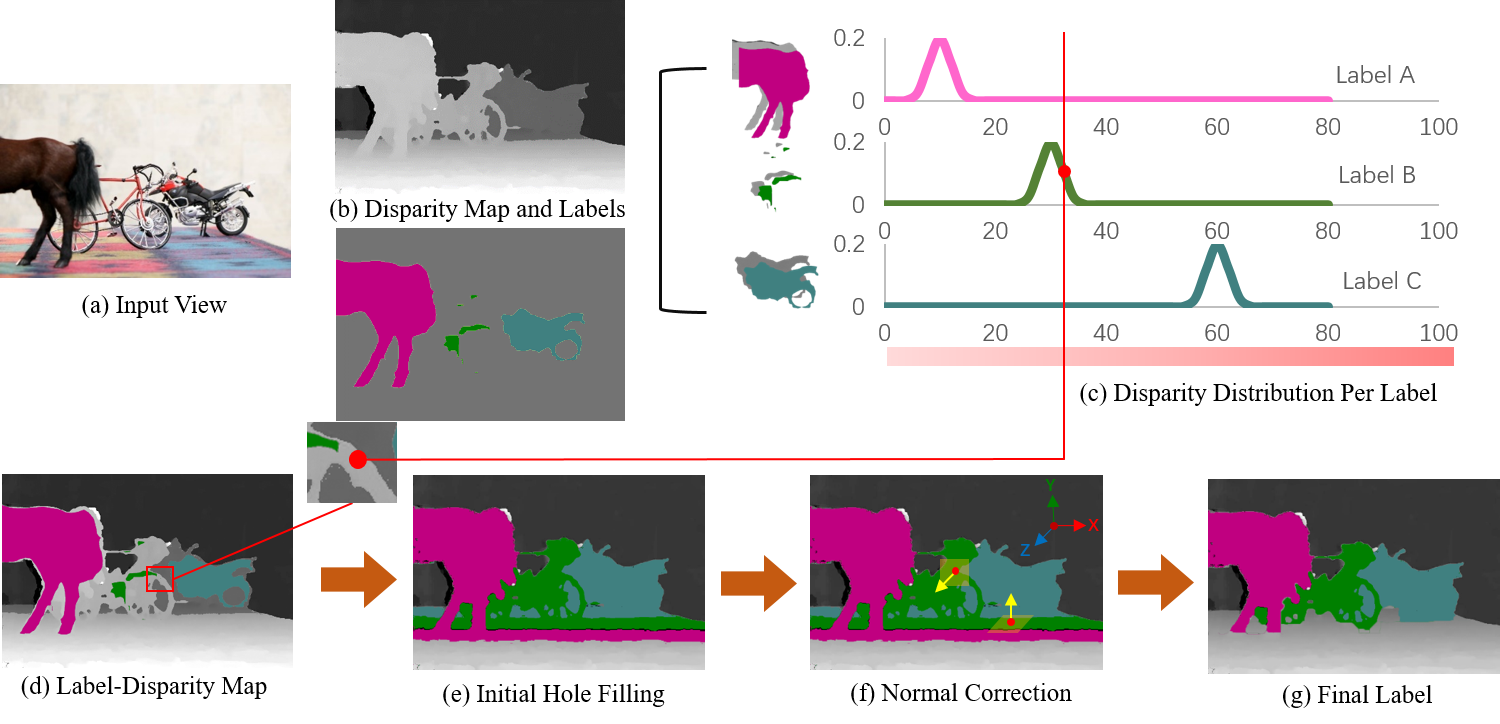}
 \caption{Depth-guided Semantic Label Refinement. Given the initial HCSM, we first model, for each label, the distribution of its disparity (c). For each unlabeled (low confidence) pixel, we query its disparity (d) into (c) to determine its optimal label. Finally, we remove the outliers caused by the ground plane via normal comparison (f).}
\label{fig.label_refinement_paradigm}
\end{figure*}

\begin{figure}
\includegraphics[width=1\linewidth]{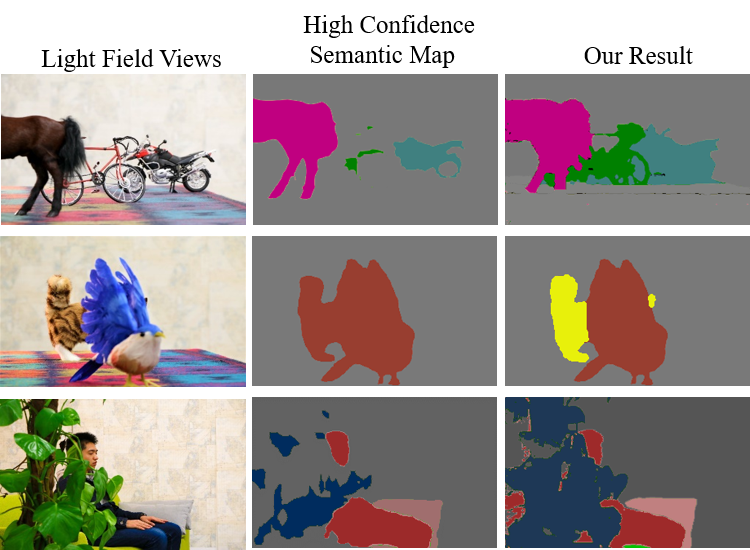}
\caption{Semantic labeling before and after stereo refinement. Left shows the
input view; Middle shows the high confidence semantic map (HCSM); Right
shows the stereo refined results.}
\label{fig.label_refinement_result}
\end{figure}

It is important to note that the assumption breaks down when there is a ground plane or two objects are close to each other. For example, in Fig.\ref{fig.label_refinement_paradigm} a bicycle, a motorcycle and a horse are positioned on the same table where the table top plane maps to continuously varying disparity of a wide range. Consequently, different parts of the table will have the same disparity as the bicycle, the motorcycle, and the horse respectively. Brute-force application of the scheme above assigns the corresponding table parts as bicycle, motorcycle and horse. 

To resolve this ambiguity, we employ a refinement process that incorporates both distance and normal cues, as shown in Fig.\ref{fig.label_refinement_paradigm}. For each conflict position $\textbf{p}$, we assume it belongs to either object A or B. We use two indicator functions $1_A(\textbf{p})$ and $1_B(\textbf{p})$ to represent that $\textbf{p}$ belongs to A or B. We use two measures to locally evaluate the possibility of each assignment for each conflict pixel. They comprise normal consistency and distance consistency,

\begin{equation}
E=p_n \cdot E_n(1_A(\textbf{p}), 1_B(\textbf{p})) + p_d \cdot E_d(1_A(\textbf{p}), 1_B(\textbf{p}))
\label{equ.semantic_disambiguity}
\end{equation}
$E_n$ represents the normal consistency and $E_d$ represents the distance consistency. 

To measure normal consistency, we extract confident regions $\mathcal{R}^{\{A, B\}}$ of A and B from HCSM, and calculate normals $n$ for each pixels in $\mathcal{R}^{\{A, B\}}$ from the disparity map. Then calculate the normal consistency as
\begin{equation}
C^{\{A, B\}}=-\frac{\sum_{r\in \mathcal{R}^{\{A, B\}}}<m_i, n(r)>}{\sum_{r\in \mathcal{R}^{\{A, B\}}}<m_i, n(r)> + \tau}
\end{equation}
where$<\cdot>$ represents inner product, $m$ if the normal of $\textbf{p}$ from the disparity map. $C^{\{A, B\}}$ will be close to 1 when the normal of $\textbf{p}$ is consistent with $\mathcal{R}^{\{A, B\}}$, i.e., parallel with normals in $\mathcal{R}^{\{A, B\}}$. This indicates that $\textbf{p}$ probably belongs to A (B). Otherwise, $\textbf{p}$ is less likely to belong to A (B). Thus, $E_n$ in Eq.\ref{equ.semantic_disambiguity} is 
\begin{equation}
E_n = C^A \cdot 1_A(\textbf{p}) + C^B \cdot 1_B(\textbf{p}) 
\end{equation}

We also calculate the distance consistency as follows:
\begin{equation}
E_d^{\{A, B\}}=\frac{\min_{\textbf{r}\in \mathcal{R}^A} \|\textbf{p}-\textbf{r}\|^2}{\max_{\textbf{r}\in \mathcal{R}^A} \|\textbf{p}-\textbf{r}\|^2} \cdot 1_A(\textbf{p})+ \frac{\min_{\textbf{r}\in \mathcal{R}^B} \|\textbf{p}-\textbf{r}\|^2}{\max_{\textbf{r}\in \mathcal{R}^B} \|\textbf{p}-\textbf{r}\|^2} \cdot 1_B(\textbf{p})
\end{equation}
$E_d$ will be close to 0 when $\textbf{p}$ is close to the region of A. Otherwise, it will be a large value.

We consider normal consistency only when one of A or B is ground region. In this case, we can easily calculate the normal of the ground region from the disparity map.

\section{Semantic See-Through Rendering}
\label{sec:weighting}

\begin{figure}
\includegraphics[width=1\linewidth]{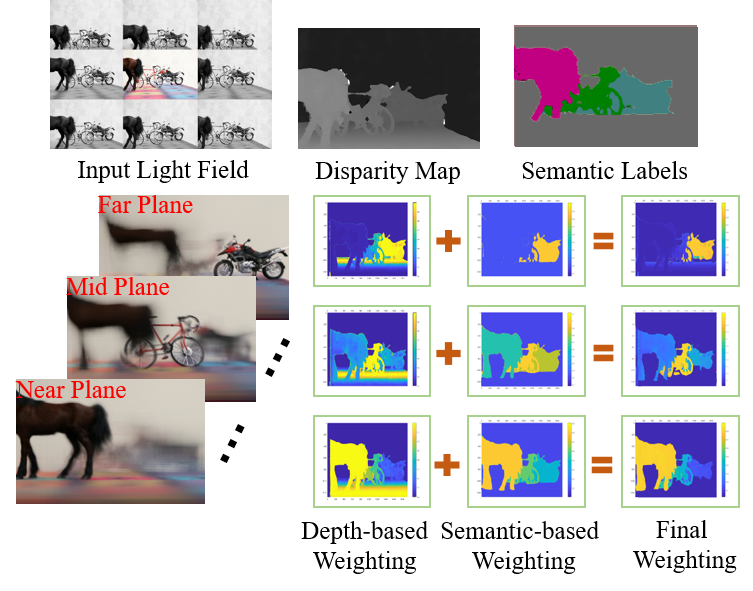}
\caption{SST Rendering: We assign every ray a weight in terms of its disparity and semantic label. The three rows show how weighting changes in terms of the focal depth.}
\label{fig.render_weight}
\end{figure}

Once we manage to assign a label and disparity to each ray in the light field, we can conduct semantic see-through (SST) rendering. We first modify the traditional LF rendering by adding a new weighting function:

\begin{equation}
\begin{split}
E^*(u', v') &= \iint W(s_r,d_r,d_f)\\
&L(s,t,s+\frac{u'-s}{d_f},t+\frac{v'-t}{d_f})A(s,t)\mathrm{d}s\mathrm{d}t
\end{split}
\end{equation}
where $r = [s,t,s+\frac{u'-s}{d_f},t+\frac{v'-t}{d_f}]$,  $W(s_r, d_r, d_f)$ is the (normalized) weighting function of both the semantic label $s_r$ and the depth $d_r$ of ray $r$. An example of out weighting scheme is shown in Fig.\ref{fig.render_weight}.

\paragraph{Depth-guided Weighting.} We design the special weighting function in order to achieve better see-through results (Fig.\ref{fig.render_weight}). Suppose a desired ray $r_d$ intersects the focal plane at point $p_{d_f}$. For each sub-aperture view $C(s,t)$, we trace the point back to $C$ as ray $r_{st}$. Since we know the corresponding disparity map for $C(s,t)$, we obtain two depth values, $d_f$ as the focal plane depth and $d_{r_{st}}$ as the depth/disparity of the ray. Conceptually if the ray's depth coincides with the focal plane, the ray should be in-focus, and hence has a higher weight. Therefore, we assign a Gaussian function as its weight:
\begin{equation}
W^*(d_{r_{st}}, d_f) = e^{-(d_f - d_{r_{st}})^2/2\sigma_{d}^2}
\end{equation}
where $\sigma_{d}$ is the standard deviation and it determines how many rays are impactive, i.e., may has bigger weights. As $\sigma_{d}$ approaches infinity, $W^*(d_{r_{st}}, d_f)$ will be $1$ and the algorithm degrades to traditional refocusing algorithm.

However, this weighting function can not satisfy our rendering demand because it will assign a extremely small weight to $r_{st}$ when $|d_f - d_{r_{st}}|>3\sigma_{d}$. This yields unpleasant rendering effect that some objects are almost invisible. To overcome this problem, we introduce a constant minimum weight threshold $C_1$ and the depth weighting function becomes
\begin{equation}
\label{eq:depth_weight}
W(d_{r_{st}}, d_f) = (1-C_1)W^*(d_{r_{st}}, d_f) + C_1
\end{equation}

$C_1$ determines a fading factor. As $C_1$ decreases, objects away from the focal depth get more impactive. By setting different values to $C_1$, we can achieve different see-through effects.

\paragraph{Semantic-guided Weighting.} Consider refocusing on an object under occlusions.  
The first issue is when an object's scale is much smaller than the impactive depth range defined by $\sigma_{d}$. In this situation, the renderer will still assign considerable weights to its obstructions. As a result, the object will still appear with ghosting effects. The second issue is when two objects are overlapping. Consider a scene where a panda is climbing a bamboo tree, from the camera's perspective, they have similar depths. Therefore, depth-guided weighting scheme will not distinguish them very well and produce expected occlusion-removed results.

To overcome the first problem, we apply a semantic weighting function (Fig.\ref{fig.render_weight}). We assume each object's weight follows the quadratic distribution and define semantic weight function as following:
\begin{equation}
W^*(s_{r_{st}}, d_f) = max\{0, -\frac{(d_f-D_{min}^{st})(d_f-D_{max}^{st})}{((D_{max}^{st}-D_{min}^{st})/2)^2}\}
\end{equation}
where $D_{min}^{st} = min\{d_r: s_r=s_{r_{st}}\}$ and $D_{max}^{st} = max\{d_r: s_r=s_{r_{st}}\}$ define the depth range of all rays that has the same label with $r_{st}$. 

Same as depth weight blending, we define a minimum weight threshold $C_2$ so that rays do not get dark when the focal plane depth is out of their depth range. The semantic weighting function then becomes
\begin{equation}
\label{eq:semantic_weight}
W(s_{r_{st}}, d_f) = (1-C_2)W^*(s_{r_{st}}, d_f)+C_2
\end{equation}
$C_2$ determines another fading factor.

In the second scenario where two objects occlude each other, it is non-trivial to tell which object is more important and should be clear. Our implementation utilizes a user interface to choose the desired one manually. We then reduce the weight of other objects by a factor of 2.

\begin{figure*}
\includegraphics[width=1\linewidth]{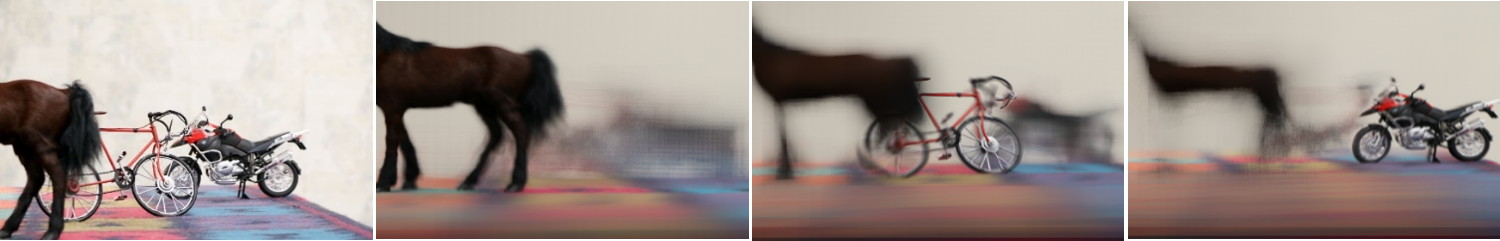}
\caption{SST rendering on the "transportation" light field composed of a toy horse, a bicycle model, and a motorcycle model.}
\label{fig.rendering_result_2}
\end{figure*}

By multiplying $W(d_{r_{st}}, d_f)$ and $W(s_{r_{st}}, d_f)$, we obtain a joint weight function:
\begin{equation}
W(s_r, d_r, d_f) = W(d_{r_{st}}, d_f)\cdot{}W(s_{r_{st}}, d_f)
\end{equation}
Normalizing over $st$ gives us the final 
\begin{equation}
W_{norm}(s_r,d_r, d_f) = \frac{W(d_{r_{st}}, d_f)\cdot{}W(s_{r_{st}}, d_f)}{\iint W(d_{r_{st}}, d_f)\cdot{}W(s_{r_{st}}, d_f)\mathrm{d}s\mathrm{d}t }
\end{equation}

The complete algorithm can be easily implemented on the GPU as in Algorithm~\ref{al:render}.

\begin{algorithm}
\caption{Real-time semantic light field rendering}\label{euclid}
\begin{algorithmic}[1]
\Require Light field $C(s,t)$; Depth map $D(s,t)$; Semantic map $S(s,t)$
\ForAll {pixel $(x,y)$ on desired image plane} 
	\State $p:=$ the pixel under consideration
    \State $q:=$ the intersection of $\overrightarrow{O_rp}$ with focal plane
    \State $W_{sum}:=0$
    \State $c(x,y):=BLACK$
    \ForAll {reference camera $(s,t)$ in aperture}
		\State $r:=$ the ray through $q$ and $O(s,t)$
        \State $(u,v):=$ projection of $q$ onto the image plane of $(s,t)$
        \State Compute $W_{depth}$ using Eq.\ref{eq:depth_weight}
        \State Compute $W_{semantic}$ using Eq.\ref{eq:semantic_weight}
        \State $W:=W_{depth} * W_{semantic}$ 
        \State $W_{sum}:=W_{sum} + W$
        \State $c(x, y):=c(x, y) + W * C(s,t,u,v)$
    \EndFor
    \State $c(x, y) := c(x,y) / W_{sum}$
\EndFor
\end{algorithmic}
\label{al:render}
\end{algorithm}

\section{Experimental Results}

To evaluate our algorithm, we set up different kinds of scenes to capture real data. The supplementary materials are also provided to show more results. We first build a mechanical camera array using a Canon 760D camera to record indoor light field data in $1920 \times 1280$ resolution for each subaperture view. We mount the camera on a motor to control its position precisely. We also build a $6 \times 6$ GoPro camera array to capture outdoor scenes. The resolution of each subaperture view is $2000 \times 1500$ and all cameras can be synchronized by the official synchronizer (Fig.\ref{fig:Teaser}). Moreover, we test our rendering scheme on Stanford light field archive~\cite{Vaish2004UsingP}. For our captured scenes, the input are only light field images. All our experiments are conducted on a computer with Intel Xeon CPU E5-1620 v4 @ 3.5GHz, 16GB memory and Titan X graphics card.

To conduct our SST algorithm, we need to compute the semantic labels and depth maps for each subaperture view first. We implement our stereo-based semantic refinement algorithm in MATLAB and compare our results with that of PSPNet -- one of the most advanced semantic segmentation approaches. Fig.\ref{fig.label_refinement_result} shows that our results exhibit much higher accuracy and sharper boundaries. Our refinement can handle heavily occluded areas which are very hard to deal with in single image semantic segmentation. As for the depth map generation, we compare our approach with MC-CNN method. Due to the occlusion existing, MC-CNN method generates large black holes especially at the boundaries of the objects as shown in the closeups of Fig.\ref{fig:depth_warp}. Instead, our approach patches mismatched and occluded positions effectively (Fig.\ref{fig:depth_warp}) by exploiting disparity consistency among adjacent views. The improvement of disparity maps also in turn enhances our semantic segmentation refinement. 

\begin{figure}
\includegraphics[width=1\linewidth]{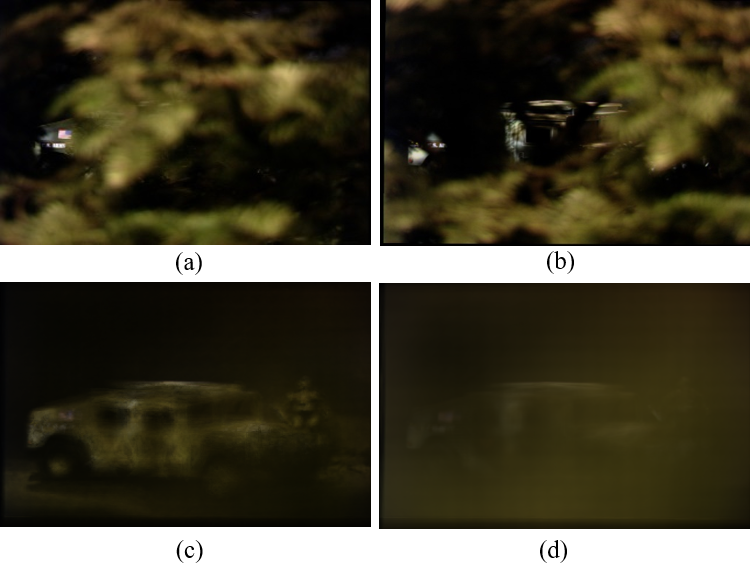}
\caption{SST rendering vs. regular rendering on the Stanford "Toy Humvee and soldier" light field. (a) and (b) are two views in the light field. Our technique (c) is able to significantly enhance the see-through capability, while regular refocusing exhibits strong foreground residue.}
\label{fig.rendering_result_3}
\end{figure}

\begin{figure}
\includegraphics[width=1\linewidth]{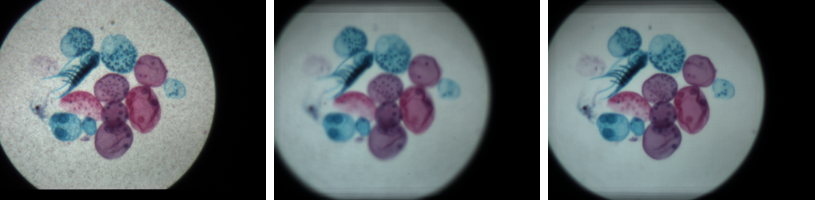}
\caption{SST rendering on the "cell" light field composed of stained cells.}
\label{fig.rendering_result_5}
\end{figure}

We implement SST on a NVIDIA GPU. Fig.\ref{fig.rendering_result_2} shows our SST rendering results on the "transportation" light field dataset captured by our Canon device. When the focus changes from the foreground to the background, different models (the toy horse, the bicycle model and the motorcycle model) become clear. Even the motorcycle is occluded by the bicycle, our SST can still render the complete motorcycle thanks to the additional semantics and depth information. Our SST method works on not only the large scale scenes, but also the micro scenes like cells. Fig.\ref{fig.rendering_result_5} shows our SST rendering results on the "cell" light field. We change the focal plane to distinguish organelles at two depth layers.
"Toy Humvee and soldier" light field from Stanford archive is composed of two focal layers, e.g.,the bushy leaves and the toys. Our rendering scheme produces a remarkable see-through effects while this is impossible by using traditional rendering scheme as shown in Fig.\ref{fig.rendering_result_3}.

\begin{figure*}
\includegraphics[width=1\textwidth]{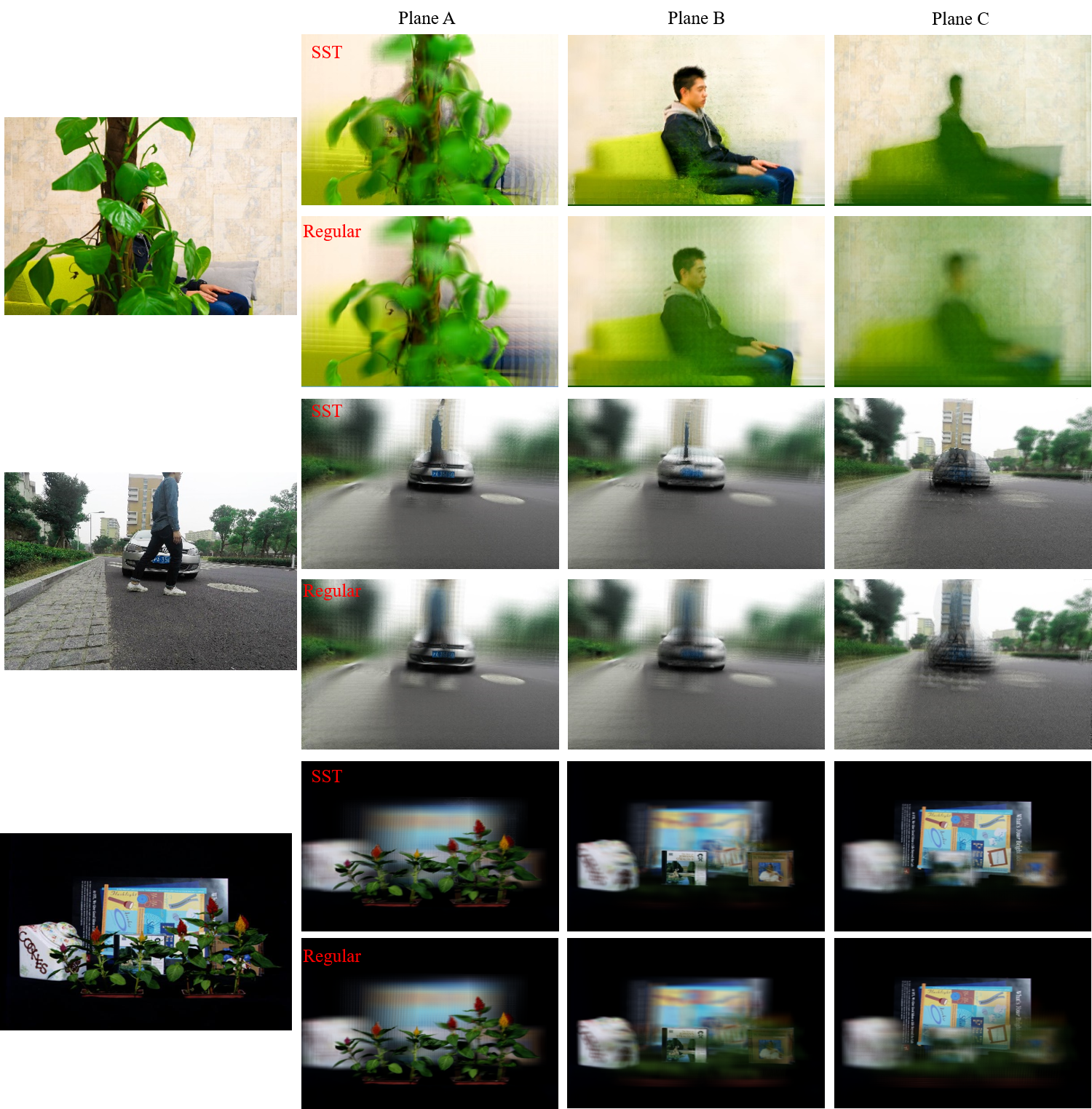}
\caption{SST Rendering vs. Regular Refocusing on the "bush" light field (top row), "pedestrian" light field (mid row), and the Standford "CD cases and poster" light field (bottom row). From left to right, we change the focal plane. Our technique achieves superb see-through capability while maintaining smooth focus transitions. Please refer to the supplementary video for the complete focal sweep results. }
\label{fig.rendering_result_1}
\end{figure*}

We also compare the rendering results of SST with traditional refocusing's on both indoor and outdoor scenes as shown in Fig.\ref{fig.rendering_result_1}. SST remarkably reduces ghosting effects and produces a more clear view of occluded objects than traditional refocusing. In the first scene, we capture the "bush" light field. A man sits on a sofa behind the heavily foliaged bush which extremely occludes him. Traditional refocusing generates obvious ghosting artifacts on the face and body, because the method doesn't take the label information of the environment into account. Instead, SST leverages the depth and labels to assign each ray with different weight and therefore remove nearly all foreground bush when the focus is on the man. Seeing through the occluders is crucial in security surveillance. The second scene is the "pedestrian" light field. A man walks in front of a white car. When focusing on the car, SST decreases the impacts of the pedestrian, and renders the car more clearly. When focusing on the building at behind, traditional scheme renders a mixed color on the building surface, while SST renders the correct color. To test the robustness of our algorithm, we also compare our SST with regular refocusing method on the Stanford "CD cases and poster" light field (the bottom row in Fig.\ref{fig.rendering_result_1}). Our SST still achieves superb see-through effects when the focus is on the CD cases and poster individually.

\begin{figure}
\includegraphics[width=1\linewidth]{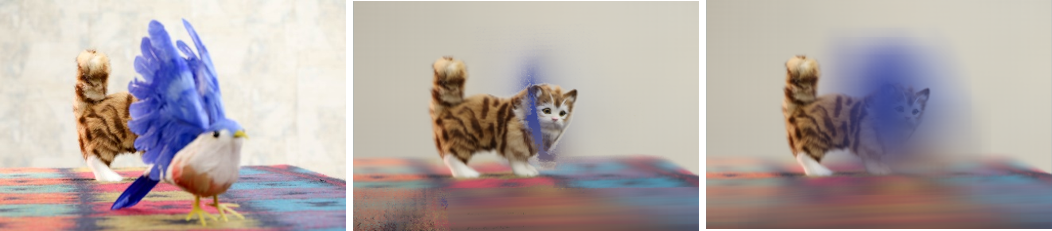}
\caption{On the "cat" light field, even SST did not manage to completely remove the foreground hen model (left), it still incurs much less residue compared with regular refocusing (right).}
\label{fig.rendering_result_4}
\end{figure}

Although our SST has achieve unprecedented see-though rendering results, it still has some drawbacks (Fig.\ref{fig.rendering_result_4}). Our SST algorithm is based on input disparity maps and semantic labeling, so the quality of the input is crucial to generate good results. SST can fail when stereo matching algorithm does not compute reasonable disparity values, and PSPNet incorrectly label regions of objects at the same time. Because in this case, we are not able to refine semantic maps and SST is unable to differentiate rays according to its label and depth.

\section{Conclusions and Future Work}
We have presented a novel semantic see-through (SST) rendering technique that can provide high quality refocusing effects. Different from traditional refocusing, our technique effectively exploits semantic ray labeling and learning-based stereo matching. By designing proper blending functions, our technique manages to mitigate visual artifacts caused by foreground residues and at the same time maintains smooth spatial, angular and focal transitions. 

To our knowledge, this is the first light field rendering technique that explicitly incorporates semantic analysis. There is also an emerging trend on integrating semantic labeling into 3D reconstruction~\cite{7575643, Ladicky2010JointOF} such as stereo matching and volumetric reconstructions. Our immediate task is to explore visibility analysis in these techniques for more general image-based rendering techniques such as view morphing~\cite{Seitz1996ViewM}, lumigraph~\cite{Gortler1996TheL}, and surface light field rendering~\cite{Wood2000SurfaceLF}. Further, more sophisticated label sets, e.g., to separate various materials, can be used in our framework. This can potentially benefit 3D reconstruction highly view-dependent materials. For example, we intend to explore coupling semantic analysis with specular light field reconstruction~\cite{wang2016svbrdf}, light field super-resolution~\cite{6239346}, and hyper-spectral light field imaging~\cite{Holloway2015GeneralizedAC}. 

\bibliographystyle{ACM-Reference-Format}
\bibliography{sample-bibliography} 

\end{document}